\documentclass[runningheads]{llncs}
\usepackage[T1]{fontenc}

\usepackage{times}
\usepackage{latexsym}
\usepackage{graphicx}
\usepackage{booktabs}
\usepackage{rotating}

\usepackage[utf8]{inputenc}

\usepackage{microtype}

\usepackage{inconsolata}
\usepackage[T1]{fontenc}
\usepackage{amsmath}
\usepackage{amssymb}

\usepackage{tcolorbox}        
\tcbuselibrary{breakable}     
\tcbuselibrary{skins}         
\tcbuselibrary{theorems}      
\usepackage{graphicx}
\usepackage[table]{xcolor} 
\usepackage{arydshln}
\usepackage{amssymb}
\usepackage{pifont}

\definecolor{bg}{rgb}{0.95,0.95,0.95}
\usepackage{listings} 

\usepackage{tabularx}
\usepackage{makecell}
\usepackage{ragged2e}
\usepackage{float}

\usepackage{booktabs}   
\usepackage{multirow}   
\usepackage{times}
\usepackage{soul}
\usepackage{url}
\usepackage[breaklinks]{hyperref}  
\usepackage[hyphenbreaks]{breakurl}
\PassOptionsToPackage{hidelinks}{hyperref}

\usepackage{booktabs}
\usepackage{algorithm}
\usepackage{algorithmic}
\usepackage{cleveref}
\usepackage{rotating}
\usepackage{booktabs}
\usepackage{longtable}
\usepackage{graphicx}

\definecolor{lightpurple}{RGB}{235,225,250}
\begin{document}
\title{Analysis of Inverse Relation Directionality in Large Language Models}
%
%
\author{Sefika Efeoglu\inst{1}\and
Adrian Paschke\inst{1,2}}

\authorrunning{S. Efeoglu and A. Paschke}

\institute{
Institute of Computer Science, Department of Mathematics and Computer Science, Freie Universität Berlin, 14195 Berlin, Germany\\
\email{sefika.efeoglu@fu-berlin.de, adrian.paschke@fu-berlin.de}
\and
Data Analytic Center (DANA), Fraunhofer Institute FOKUS, 10589 Berlin, Germany
}

\maketitle              
\begin{abstract}
Large language models (LLMs) have achieved strong performance on text-to-knowledge graph generation and related tasks. Nevertheless, it is still unclear whether they accurately model the direction-dependent semantics of inverse relations, in which reversing the order of the arguments alters the meaning of a relation (e.g., \textit{mother} versus \textit{child}). To the best of our knowledge, this work presents the first systematic study of inverse relation directionality in LLMs, using a benchmark consisting of 5,457 instances spanning 27 distinct inverse relation labels. We evaluate five open-source LLMs under a multiple-choice prompting framework and further examine the influence of relation descriptions and entity representations by substituting the original entities with synthetic and masked entities. Our findings reveal systematic asymmetries in inverse relation classification across LLMs, indicate that relation descriptions do not consistently improve performance, and show that model performance can be sensitive to variations in entity representations.

\end{abstract}
\section{Introduction}
\label{sec:intro}

Large language models (LLMs) are increasingly applied to structured knowledge acquisition tasks, including ontology learning, ontology matching, text-to-knowledge graph (KG) generation, and relation extraction~\cite{KOUTSIANA2026100873}. Since KGs are labeled directed graphs, accurately modeling the directionality of relations is essential for KG completion~\cite{Yu2023}. Relations encode semantic dependencies between entities, and their meaning depends not only on the participating entities but also on the direction of the relation~\cite{alt-etal-2020-probing}. Consequently, correctly identifying relation direction improves both KG completeness~\cite{janna_2023} and the semantic accuracy of extracted knowledge.

Inverse relations, also known as converse relations, represent the reverse of a binary relation and should not be confused with symmetric relations. For instance, the triple \textit{(x, part of, y)} corresponds to the inverse triple \textit{(y, has part, x)}. Previous studies have shown that relation classification models struggle when inverse relation pairs are included in the label space~\cite{ijcai2022p407,kemit_2025,Wu2023,Yu2023}. Nonetheless, existing evaluations remain limited. \cite{qi-etal-2023-investigation} investigate converse relations in LLMs but evaluate relation understanding at the triple level rather than sentence-level relation classification, while \cite{semaval_inverse} study direction-sensitive variants of the same relation label rather than explicit inverse relation pairs. We distinguish between these two settings: (i) true inverse (converse) relations (e.g., \textit{parent}/\textit{child}) and (ii) direction-sensitive variants of the same relation label (e.g., \textit{Member-Collection(e1,e2)} versus \textit{Member-Collection(e2,e1)}). To the best of our knowledge, no previous work has systematically evaluated whether LLMs correctly recognize inverse relation directionality at the sentence level~\footnote{The code and data used in this work will be shared upon acceptance}.

\begin{figure}[H]
    \centering
    \includegraphics[width=\linewidth]{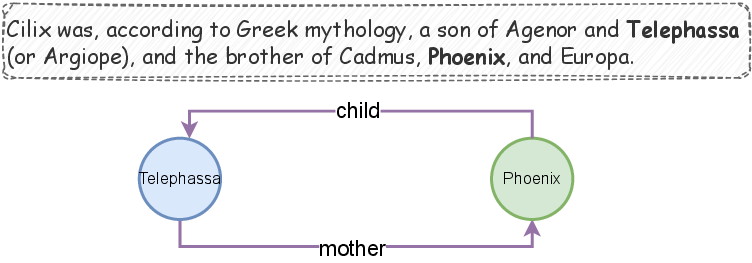}
    \caption{Example of an inverse relation pair: \textit{child} versus \textit{mother}.}
    \label{fig:inverse_pair}
\end{figure}

In this work, we address the following research question: \textit{Do instruction-tuned LLMs correctly recognize inverse relation directionality?} To answer this question, we conduct a systematic analysis of inverse relation directionality in LLMs. We evaluate whether instruction-tuned LLMs correctly recognize inverse relation pairs in both head-to-tail and tail-to-head directions using a benchmark comprising 27 inverse relation labels derived from FewRel~\cite{han-etal-2018-fewrel} and TekGen~\cite{text_to_kg}. To analyze the effect of entity familiarity, we further evaluate the same LLMs after replacing the original entities with synthetic or masked entities, thereby separating relational reasoning from potential familiarity with real-world entities. Our contributions are summarized as follows:

\begin{itemize}
    \item We present the first systematic analysis of inverse relation directionality in LLMs for sentence-level relation classification.
    \item We introduce a benchmark of inverse relation pairs derived from FewRel and TekGen, with direction-verified labels for head-to-tail and tail-to-head relation classification using LLMs.
    \item We investigate the effects of relation descriptions and entity familiarity through controlled entity perturbation experiments.
\end{itemize}

\noindent The remainder of this paper is organized as follows. \Cref{sec:preliminaries} formally introduces inverse relations. \Cref{sec:related_works} reviews related work. \Cref{sec:methodology} describes the benchmark construction and evaluation methodology. \Cref{sec:evaluation} presents the experimental results, while \Cref{sec:discussion} discusses their implications. Finally, \Cref{sec:conclusion} presents the concluding remarks in this work.
\section{Preliminaries}
\label{sec:preliminaries}

Inverse relations, also known as converse relations, denote the reverse of a binary relation. We formally define the inverse of a binary relation in \Cref{eq:ch_4_inverse_relation_def} and \Cref{eq:ch_4_inverse_relation_eq}. The assumptions adopted throughout this work are summarized in the definition below.

\begin{tcolorbox}[
    title={Inverse Relation Definition},
    breakable
]
Let $R$ be a binary relation on a domain $D$. The \emph{inverse} of $R$, denoted by $R^{-1}$, is the binary relation on $D$ defined by
\begin{equation}
R^{-1}xy \;\iff\; Ryx
\quad\text{for all } x,y \in D.
\label{eq:ch_4_inverse_relation_def}
\end{equation}

Equivalently,
\begin{equation}
R^{-1}
=
\{(x,y)\in D\times D \mid (y,x)\in R\}.
\label{eq:ch_4_inverse_relation_eq}
\end{equation}

\paragraph{For the purposes of this work, we make the following assumptions:}
\begin{itemize}
    \item $R$ or $R^{-1}$ may correspond to gender-specific subrelations of a more general relation. For example,
    $\mathrm{Wife}, \mathrm{Husband} \subseteq \mathrm{Spouse}$ and
    $\mathrm{Mother}, \mathrm{Father} \subseteq \mathrm{Parent}$.

    \item If $R$ or $R^{-1}$ is one of the subrelations of a more general relation, it cannot simultaneously hold for the same ordered pair $(x,y)$.
\end{itemize}
\end{tcolorbox}

In this work, inverse relations are evaluated at the level of KG properties and their sentence-level interpretation rather than according to strict logical equivalence for every possible subrelation pair.
\section{Related Work}
\label{sec:related_works}
\noindent This section summarizes previous work on relation classification, with a particular focus on inverse relations.

Regarding datasets, \cite{parekh_2020} analyze benchmark datasets derived from Wikidata and DBpedia, noting that they contain various types of canonical relations, including hierarchical, inverse, and ownership relations. Their work primarily examines relations among \textit{Organization}, \textit{Person}, and \textit{Location} entities and further explores canonical relation patterns in the FewRel dataset. However, they do not evaluate the difficulty of relation classification. \cite{qi-etal-2023-investigation} investigate whether LLMs can understand the semantics of converse relations within sentences. They introduce the ConvRE dataset, which consists of multiple-choice questions formulated as Re2Text and Text2Re tasks. They focus on semantic understanding rather than sentence-level relation classification and exclude symmetric relations from their analysis. Similarly, \cite{semaval_inverse} examine whether relation classification models can learn directional semantics using the SemEval dataset. However, this dataset does not explicitly include inverse relation pairs; instead, it contains direction-sensitive variants of the same relation.

Several works report performance degradation in relation classification models when the predefined label space includes inverse relation pairs, particularly in few-shot settings~\cite{ijcai2022p407,kemit_2025,Wu2023,Yu2023}. \cite{ijcai2022p407} demonstrate that inverse relation classification remains challenging in few-shot scenarios and propose a phrase-level functional word encoder to address this issue. Their approach leverages class-general attention to model the importance of function words. \cite{Wu2023} introduce an entity- and relation-centered encoding method to improve inverse relation recognition in FewRel 1.0 under few-shot settings. \cite{Yu2023} propose a block-decomposition-based relation interaction framework for temporal KG completion that incorporates inverse relations to enhance modeling. Similarly, \cite{kemit_2025} construct symmetric graphs from inverse relations, employ LLMs to generate relation descriptions, and compare their approach with text-based methods for KG completion. Collectively, these studies emphasize the importance of accurately modeling inverse relations and consistently report performance declines when inverse relation pairs are included in the label space.

Additionally, \cite{lippolis_2025} evaluate the ontology generation capabilities of LLMs, including their ability to recognize inverse relations (\textit{inverseOf}). Their results show that LLMs struggle to correctly identify inverse relations. Similarly, \cite{Fathallah_2026} apply LLMs to ontology learning and report comparable difficulties in recognizing inverse relations. However, both studies evaluate ontology generation using ontology benchmarks rather than sentence-level relation classification benchmarks, which are the focus of this work.

In summary, previous works have not systematically evaluated inverse relation directionality in sentence-level relation classification using LLMs. Our work enables such an empirical analysis.
\section{Methodology}
\label{sec:methodology}
\begin{figure}[H]
    \centering
    \includegraphics[width=0.8\linewidth]{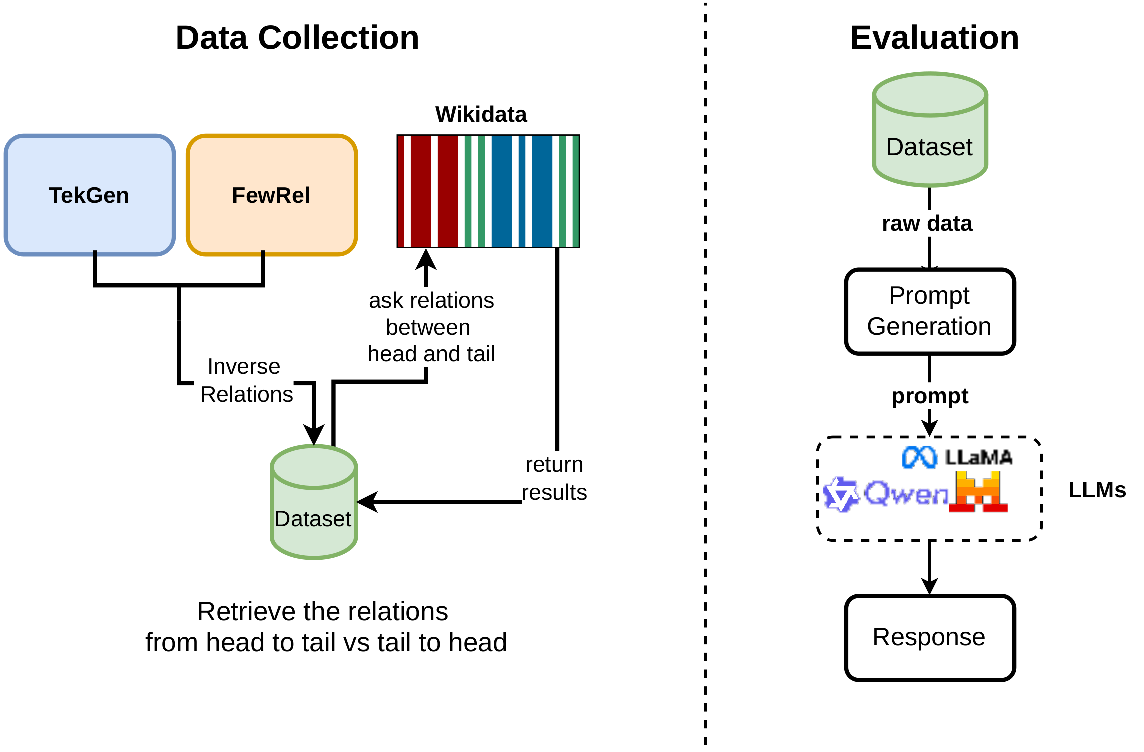}
    \caption{Overview of the benchmark construction and evaluation pipeline.}
    \label{fig:methodology}
\end{figure}

This section describes the benchmark construction and evaluation methodology used to systematically analyze inverse relation directionality in LLMs. We first present the benchmark construction process and the selection of inverse relation pairs in \Cref{sec:dataprep}. Next, we introduce the zero-shot multiple-choice prompting strategies in \Cref{sec:prompting}. Finally, \Cref{fig:methodology} provides an overview of the complete pipeline, including benchmark construction from FewRel~\cite{han-etal-2018-fewrel} and TekGen~\cite{text_to_kg}, prompt generation, and evaluation across multiple LLMs.

\subsection{Data Collection}
\label{sec:dataprep}

To enable a systematic evaluation of inverse relation directionality, we construct a benchmark from FewRel 1.0~\cite{han-etal-2018-fewrel} and the TekGen partition of TEXT2KGBENCH~\cite{text_to_kg}. The benchmark construction consists of five steps: (1) selecting candidate inverse relation pairs from FewRel and TekGen, (2) extracting head and tail entity pairs, (3) querying Wikidata~\cite{wikidata_2014} for all properties connecting each entity pair, (4) filtering the retrieved properties to predefined inverse relation pairs, and (5) assigning direction-specific labels (head-to-tail and tail-to-head). Since inverse relations are direction-dependent, reversing the entity order changes the semantic interpretation of the relation.
\begin{tcolorbox}[title={SPARQL Query for Retrieving Relations}]
\begin{lstlisting}[
language=SPARQL,
basicstyle=\ttfamily\scriptsize,
numbers=left,
numbersep=6pt,
breaklines=true,
frame=none
]
SELECT ?prop ?from ?to
WHERE {
  VALUES (?from ?to) {
    (wd:QID_head wd:QID_tail)
  }
  {
    ?from ?prop ?to .
  }
  UNION
  {
    ?to ?prop ?from .
  }
  FILTER(STRSTARTS(STR(?prop), STR(wdt:)))
}
\end{lstlisting}
\end{tcolorbox}

We retrieve relation properties from Wikidata using the SPARQL query shown above. For each sentence, one relation is assigned to the head-to-tail direction and its corresponding inverse relation is assigned to the tail-to-head direction.

\begin{figure}[H]
    \centering
    \includegraphics[width=0.8\linewidth]{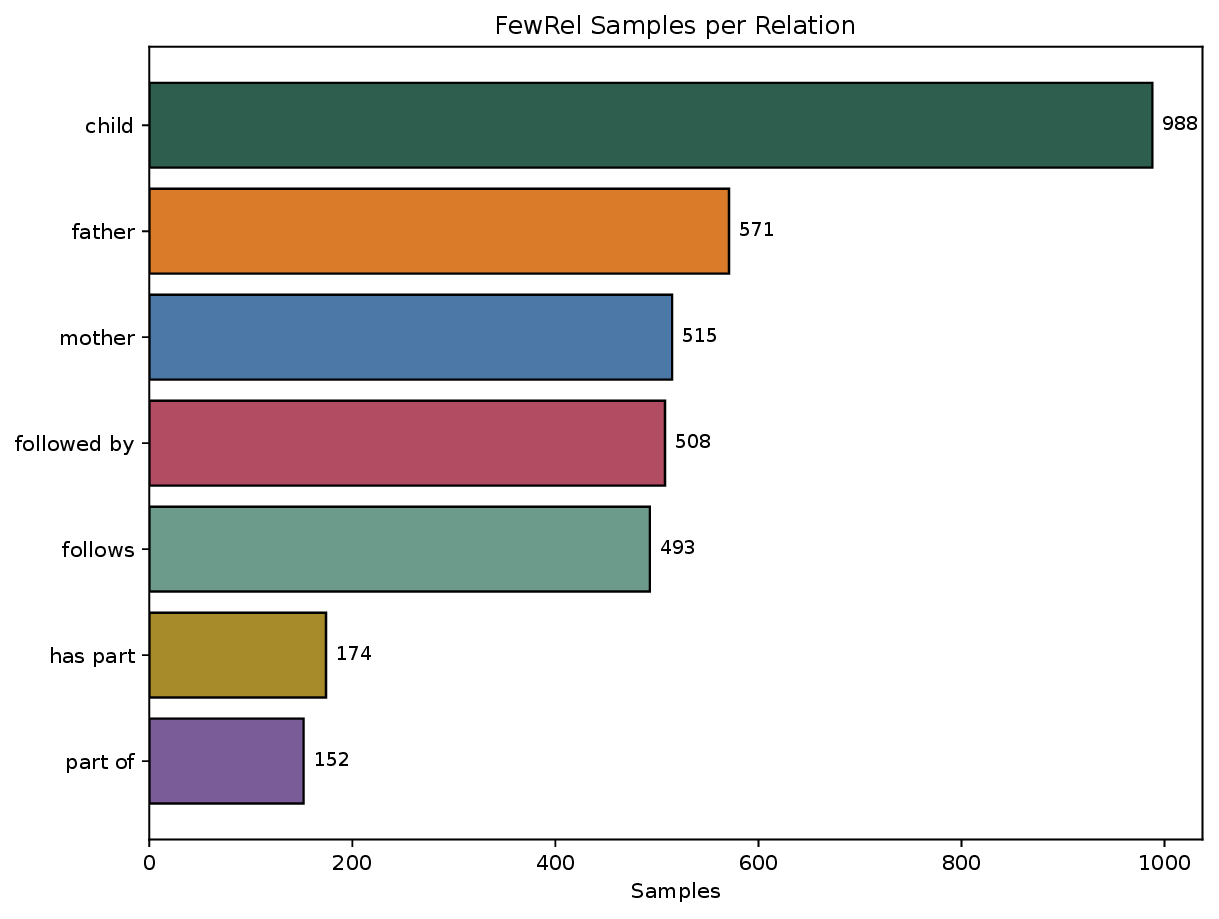}
    \caption{Statistics of head-to-tail relation labels in the dataset constructed from FewRel.}
    \label{fig:data_stats_fewrel}
\end{figure}

\begin{figure}[H]
    \centering
    \includegraphics[width=0.8\linewidth]{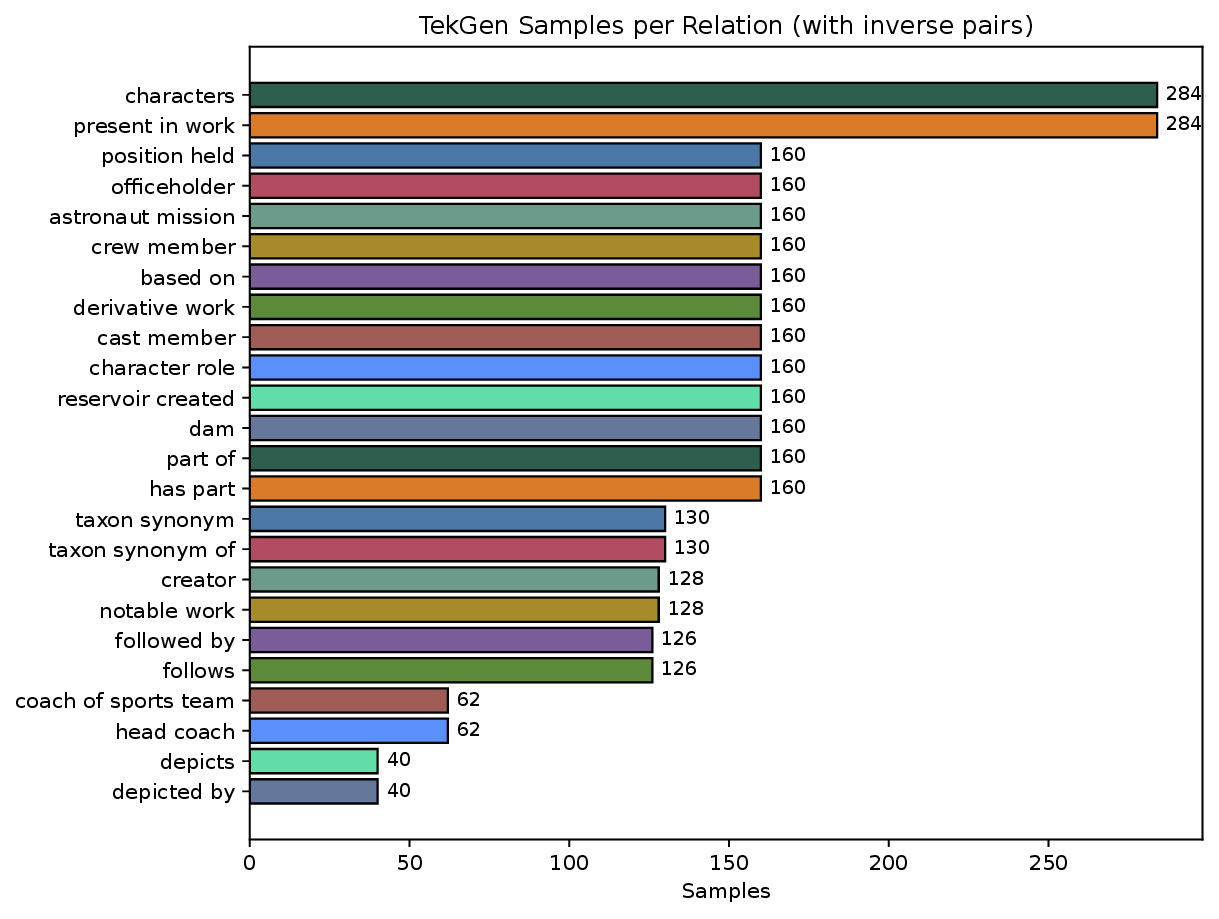}
    \caption{Statistics of head-to-tail relation labels in the dataset constructed from the TekGen partition.}
    \label{fig:data_stats_tekgen}
\end{figure}

The resulting benchmark contains 5,457 sentence-level instances, including 988 \textit{child} relation examples (\Cref{fig:data_stats_fewrel,fig:data_stats_tekgen}). Only 12 of the 16 inverse relation pairs in TekGen have associated sentence instances (see ~\Cref{fig:data_stats_tekgen} and Appendix C). Relation directions are verified against Wikidata, and entity pairs with inconsistent directional information are discarded before assigning the final labels. Consequently, all benchmark statistics reported in \Cref{fig:data_stats_fewrel,fig:data_stats_tekgen} are computed from direction-verified head and tail entity pairs obtained from FewRel and TekGen. An example from FewRel is shown in Appendix B.

\subsection{Prompting Strategies}
\label{sec:prompting}
We evaluate LLMs using zero-shot multiple-choice prompting with and without relation descriptions (see~\Cref{tab:prompts}). Each prompt contains three candidate relations: (A) the head-to-tail relation, (B) its inverse (tail-to-head), and (C) a negative relation that does not belong to the evaluated inverse relation pair. For each inverse relation pair, the negative relation is randomly selected from the remaining inverse relation labels using a fixed random seed (42). For TekGen, negative relations are sampled from all 16 inverse relation pairs listed in Appendix C, including pairs without sentence instances. The third option allows us to identify cases where the model incorrectly predicts one of the inverse relations when neither is appropriate. To analyze the effect of entity familiarity, we additionally evaluate the prompts using synthetic entities and fully masked entities. The complete evaluation settings are summarized in \Cref{tab:prompt_types_mcq}.

\begin{table}[H]
\centering
\renewcommand{\arraystretch}{1.2}

\begin{tabularx}{\textwidth}{p{2.8cm} X X}
\toprule
\textbf{Evaluation Strategy} 
& \textbf{Base Entity Tasks} 
& \makecell{\textbf{Synthetic Tasks}\\\textbf{(Anti-Memorization Tasks)}} \\
\midrule

\textbf{Without Relation}
& MCQ: Inverse relations\newline (Head-to-Tail or Tail-to-Head)
& MCQ: Inverse relations\newline (i) synthetic head/tail entities\newline (ii) masked entities \\

\hdashline

\textbf{With Relation}
& MCQ: Inverse relations with relation description
& MCQ: Inverse relations with relation description,\newline (i) synthetic head/tail entities,\newline (ii) masked entities \\

\bottomrule
\end{tabularx}
\caption{Prompt types with multiple-choice questions (MCQ).}
\label{tab:prompt_types_mcq}
\end{table}
\begin{table}[H]

\scriptsize
\centering
\renewcommand\arraystretch{1.2}
\setlength{\tabcolsep}{6pt}
\begin{tabularx}{\textwidth}{@{} >{\raggedright\arraybackslash}p{3cm} X l @{}}

\toprule
\textbf{Template} & \textbf{Question Text} & \textbf{Ground Truth} \\
\midrule

\textit{(Original) head to tail} &
\RaggedRight What is the relation from \textbf{Aage} to \textbf{Niels Bohr} in the sentence?  
Sentence: Niels Bohr and his son Aage, a physicist who acted as his father's assistant, arrived on 30 December on the first of several visits as a consultant.  
A.) child. \quad B.) father. \quad C.) followed by
Please choose A, B, or C. Answer: &
child \\

\rowcolor{green!15}
\textit{(Original) tail to head} &
\RaggedRight What is the relation from \textbf{Niels Bohr} to \textbf{Aage} in the sentence?  
\textit{Same sentence and answer options as above.} &
father \\

\hdashline

\textit{Synthetic Entities (head to tail)} &
\RaggedRight What is the relation from \textbf{Aage} to \textbf{Devin Rodriguez} in the sentence?  
Sentence: Devin Rodriguez and his son Aage, a physicist who acted as his father's assistant, arrived on 30 December on the first of several visits as a consultant.  
A.) child. \quad B.) father. \quad C.) followed by
Please choose A, B, or C. Answer: &
child \\

\rowcolor{green!15}
\textit{Synthetic Entities (tail to head)} &
\RaggedRight What is the relation from \textbf{Devin Rodriguez} to \textbf{Aage} in the sentence?  
\textit{Same sentence and answer options as above.} &
father \\

\hdashline

\textit{Masked Entities (head to tail)} &
\RaggedRight What is the relation from \textbf{XXX} to \textbf{YYY} in the sentence?  
Sentence: YYY and his son XXX, a physicist who acted as his father's assistant, arrived on 30 December on the first of several visits as a consultant.  
A.) child. \quad B.) father. \quad C.) followed by
Please choose A, B, or C. Answer: &
child \\

\rowcolor{green!15}
\textit{Masked Entities (tail to head)} &
\RaggedRight What is the relation from \textbf{YYY} to \textbf{XXX} in the sentence?  
\textit{Same sentence and answer options as above.} &
father \\

\hdashline

\textit{(Original) head to tail with relation description} &
\RaggedRight What is the relation from \textbf{Aage} to \textbf{Niels Bohr} in the sentence?  
Sentence: Niels Bohr and his son Aage, a physicist who acted as his father's assistant, arrived on 30 December on the first of several visits as a consultant.  
A.) child: subject has object as biological, foster, and/or adoptive child.  
B.) father: male parent of the subject. For stepfather, use ``stepparent'' (P3448).  
C.) followed by: immediately following item in a series of which the subject is a part [if the subject has been replaced, e.g., political offices, use ``replaced by'' (P1366]
Please choose A, B, or C. Answer: &
child \\

\rowcolor{green!15}
\textit{(Original) tail to head with relation description} &
\RaggedRight What is the relation from \textbf{Niels Bohr} to \textbf{Aage} in the sentence?  
\textit{Same sentence and answer options as above.} &
father \\

\hdashline
& \textbf{The remaining samples follow the same template with synthetic and masked entities.} &
\\

\bottomrule

\end{tabularx}
\caption{Illustration of representative examples and ground truths for inverse relations in multiple-choice questions. Entities are shown in \textbf{bold}.}
\label{tab:prompts}
\end{table}
The remainder explains the details of our zero-shot multiple-choice prompting strategies.

\subsubsection*{a.) Without and With Relation Description}

Relation descriptions are widely used in representation-based relation classification to provide semantic definitions of relation labels. We therefore evaluate whether supplying these descriptions improves inverse relation classification under zero-shot prompting.

\subsubsection*{b.) Synthetic Entity Generation}

We replace the original entities with synthetic ones to evaluate the effect of entity familiarity in LLMs. To generate synthetic entities from the original ones, we use the Presidio Anonymizer\footnote{\url{https://github.com/microsoft/presidio}, accessed on July~19,~2026.}. This tool replaces original entities with synthetic counterparts. For example, \emph{Niels Bohr} may be replaced with
\emph{Devin Rodriguez}, and \emph{Sabiha Gökçen} with
\emph{Angela Bradley}. Since Presidio cannot anonymize every entity, some names remain unchanged. To ensure complete anonymization, we additionally evaluate a fully masked setting, as described below.

\subsubsection*{c.) Entity Masking}

To eliminate all lexical information associated with named entities, we replace entity mentions with masked entities (e.g., \texttt{XXX} and \texttt{YYY}). Unlike synthetic replacement, this setting removes all entity-specific cues and enables the evaluation of inverse relation reasoning independently of entity identity.

Overall, these prompting strategies enable us to analyze three complementary aspects of inverse relation understanding in LLMs: (i) the effect of relation descriptions, (ii) sensitivity to entity familiarity through synthetic replacement, and (iii) robustness to complete entity anonymization through masked entities.
\section{Evaluation}
\label{sec:evaluation}
\noindent This section introduces the experimental setup in \Cref{sec:experimental_setup} and then presents the experimental results on our inverse relation dataset in \Cref{sec:results}.

\subsection{Experimental Setup}
\label{sec:experimental_setup}

\noindent In this section, we provide the experimental details, including the evaluated LLMs, prompting methods, and evaluation metrics. The dataset is intended primarily as an evaluation benchmark rather than a supervised training corpus; therefore, all experiments are conducted in a zero-shot setting.

\begin{itemize}
    \item \textbf{Large Language Models.}
    We evaluate five LLMs: Flan-T5 XL~\cite{Flan_T5}, Qwen2.5-7B-Instruct~\cite{qwen2025qwen25technicalreport}, Qwen3-4B-Instruct-2507~\cite{yang2025qwen3technicalreport}, Llama-3.1-8B-Instruct~\cite{grattafiori2024llama3herdmodels}, and Mistral-7B-Instruct-v0.3~\cite{jiang2023mistral} on our dataset. We use these five instruction-tuned models to examine whether they can correctly identify relation directionality (from head to tail or vice versa). We use greedy decoding for all experiments, and the models are obtained from Hugging Face\footnote{\url{https://huggingface.co/}, accessed on August~3,~2026.}.

    \item \textbf{Prompting Methods.}
    Our base prompting strategy uses a zero-shot multiple-choice question format, which can be considered a variant of zero-shot prompting, as shown in \Cref{tab:prompt_types_mcq}.~\Cref{tab:prompts} illustrates representative prompt templates prior to answer-option randomization. Each prompt contains three answer options: the target relation, its inverse relation, and a negative relation (i.e., an unrelated relation). For TekGen, negative relations are sampled from the complete set of 16 inverse relation pairs listed in Appendix C, including those without sentence instances. The order of the answer options is randomized during prompt generation. As described in \Cref{sec:prompting}, the negative relation is randomly selected from the remaining inverse relation labels using a fixed random seed (42). Furthermore, we evaluate multiple-choice questions under two prompting strategies: without and with relation descriptions. We then replace the original entities with synthetic entities generated by the Presidio Anonymizer to analyze the effect of entity familiarity on LLM performance and repeat the experiments using these synthetic entities. Finally, we completely mask the entities using variables; for example, \textit{XXX} for the head entity and \textit{YYY} for the tail entity.

    \item \textbf{Metrics.}
    We report macro-F1 because the dataset is imbalanced.

    \item \textbf{Hardware.}
    All experiments are conducted on an A100 GPU using Google Colab Pro.
\end{itemize}
\subsection{Results}
\label{sec:results}
\noindent This section presents experimental results for inverse relations from FewRel and TekGen, obtained using five LLMs. We evaluate the LLMs under two settings: (1) a multiple-choice question format without relation descriptions and (2) a multiple-choice question format with relation descriptions, each in two directions: head-to-tail and tail-to-head. Furthermore, we repeat these experiments using synthetic and masked entities.

\begin{table}[H]
  \centering

  \setlength{\tabcolsep}{5pt}
  \renewcommand{\arraystretch}{1.08}
  \resizebox{\textwidth}{!}{%
  \begin{tabular}{llllccccc}
    \toprule
    \textbf{Direction} &
    \textbf{Dataset} &
    \textbf{Description} &
    \textbf{Entity} &
    \textbf{Llama-3.1-8B} &
    \textbf{Mistral-7B} &
    \textbf{Qwen2.5-7B} &
    \textbf{Qwen3-4B} &
    \textbf{Flan-T5 XL} \\
    \midrule

    \multirow{12}{*}{\textbf{HT}}
    & \multirow{6}{*}{\textbf{FewRel}}
    & \multirow{3}{*}{\textbf{With}}
    & \cellcolor{green!20}Original
    & \cellcolor{green!20}\textbf{47.30\%}
    & \cellcolor{green!20}14.35\%
    & \cellcolor{green!20}24.91\%
    & \cellcolor{green!20}32.00\%
    & \cellcolor{green!20}40.24\% \\

    & & & Synthetic
    & \textbf{46.05\%}{\scriptsize\,($\downarrow$1.25)}
    & 5.27\%{\scriptsize\,($\downarrow$9.08)}
    & 3.76\%{\scriptsize\,($\downarrow$21.15)}
    & 27.82\%{\scriptsize\,($\downarrow$4.18)}
    & 34.93\%{\scriptsize\,($\downarrow$5.31)} \\

    & & & Masked
    & 34.58\%{\scriptsize\,($\downarrow$12.72)}
    & 11.43\%{\scriptsize\,($\downarrow$2.92)}
    & 13.14\%{\scriptsize\,($\downarrow$11.77)}
    & 39.83\%{\scriptsize\,($\uparrow$7.83)}
    & \textbf{49.28\%}{\scriptsize\,($\uparrow$9.04)} \\

    \cmidrule(lr){3-9}

    & &
    \multirow{3}{*}{\textbf{Without}}
    & \cellcolor{green!20}Original
    & \cellcolor{green!20}24.71\%
    & \cellcolor{green!20}2.27\%
    & \cellcolor{green!20}0.92\%
    & \cellcolor{green!20}\textbf{37.66\%}
    & \cellcolor{green!20}36.61\% \\

    & & & Synthetic
    & 25.54\%{\scriptsize\,($\uparrow$0.83)}
    & 2.60\%{\scriptsize\,($\uparrow$0.33)}
    & 1.11\%{\scriptsize\,($\uparrow$0.19)}
    & 34.29\%{\scriptsize\,($\downarrow$3.37)}
    & \textbf{35.25\%}{\scriptsize\,($\downarrow$1.36)} \\

    & & & Masked
    & 20.73\%{\scriptsize\,($\downarrow$3.98)}
    & 6.74\%{\scriptsize\,($\uparrow$4.47)}
    & 1.49\%{\scriptsize\,($\uparrow$0.57)}
    & 31.15\%{\scriptsize\,($\downarrow$6.51)}
    & \textbf{33.42\%}{\scriptsize\,($\downarrow$3.19)} \\

    \cmidrule(lr){2-9}

    & \multirow{6}{*}{\textbf{TekGen}}
    & \multirow{3}{*}{\textbf{With}}
    & \cellcolor{green!20}Original
    & \cellcolor{green!20}29.06\%
    & \cellcolor{green!20}25.55\%
    & \cellcolor{green!20}22.45\%
    & \cellcolor{green!20}18.79\%
    & \cellcolor{green!20}\textbf{43.85\%} \\

    & & & Synthetic
    & 29.77\%{\scriptsize\,($\uparrow$0.71)}
    & 20.06\%{\scriptsize\,($\downarrow$5.49)}
    & 23.05\%{\scriptsize\,($\uparrow$0.60)}
    & 18.45\%{\scriptsize\,($\downarrow$0.34)}
    & \textbf{37.38\%}{\scriptsize\,($\downarrow$6.47)} \\

    & & & Masked
    & 28.31\%{\scriptsize\,($\downarrow$0.75)}
    & 30.28\%{\scriptsize\,($\uparrow$4.73)}
    & 16.86\%{\scriptsize\,($\downarrow$5.59)}
    & 26.51\%{\scriptsize\,($\uparrow$7.72)}
    & \textbf{46.46\%}{\scriptsize\,($\uparrow$2.61)} \\

    \cmidrule(lr){3-9}

    & &
    \multirow{3}{*}{\textbf{Without}}
    & \cellcolor{green!20}Original
    & \cellcolor{green!20}21.14\%
    & \cellcolor{green!20}7.52\%
    & \cellcolor{green!20}12.41\%
    & \cellcolor{green!20}28.73\%
    & \cellcolor{green!20}\textbf{38.83\%} \\

    & & & Synthetic
    & 21.88\%{\scriptsize\,($\uparrow$0.74)}
    & 8.83\%{\scriptsize\,($\uparrow$1.31)}
    & 11.86\%{\scriptsize\,($\downarrow$0.55)}
    & 27.79\%{\scriptsize\,($\downarrow$0.94)}
    & \textbf{36.46\%}{\scriptsize\,($\downarrow$2.37)} \\

    & & & Masked
    & 21.78\%{\scriptsize\,($\uparrow$0.64)}
    & 14.92\%{\scriptsize\,($\uparrow$7.40)}
    & 11.01\%{\scriptsize\,($\downarrow$1.40)}
    & 30.30\%{\scriptsize\,($\uparrow$1.57)}
    & \textbf{38.62\%}{\scriptsize\,($\downarrow$0.21)} \\

    \hdashline

    \multirow{12}{*}{\textbf{TH}}
    & \multirow{6}{*}{\textbf{FewRel}}
    & \multirow{3}{*}{\textbf{With}}
    & \cellcolor{green!20}Original
    & \cellcolor{green!20}50.49\%
    & \cellcolor{green!20}11.82\%
    & \cellcolor{green!20}\textbf{69.56\%}
    & \cellcolor{green!20}67.50\%
    & \cellcolor{green!20}48.54\% \\

    & & & Synthetic
    & 47.88\%{\scriptsize\,($\downarrow$2.61)}
    & 2.94\%{\scriptsize\,($\downarrow$8.88)}
    & 14.38\%{\scriptsize\,($\downarrow$55.18)}
    & \textbf{58.60\%}{\scriptsize\,($\downarrow$8.90)}
    & 40.35\%{\scriptsize\,($\downarrow$8.19)} \\

    & & & Masked
    & 48.69\%{\scriptsize\,($\downarrow$1.80)}
    & 4.97\%{\scriptsize\,($\downarrow$6.85)}
    & 53.36\%{\scriptsize\,($\downarrow$16.20)}
    & \textbf{54.48\%}{\scriptsize\,($\downarrow$13.02)}
    & 35.35\%{\scriptsize\,($\downarrow$13.19)} \\

    \cmidrule(lr){3-9}

    & &
    \multirow{3}{*}{\textbf{Without}}
    & \cellcolor{green!20}Original
    & \cellcolor{green!20}\textbf{67.28\%}
    & \cellcolor{green!20}2.03\%
    & \cellcolor{green!20}2.72\%
    & \cellcolor{green!20}52.87\%
    & \cellcolor{green!20}47.65\% \\

    & & & Synthetic
    & \textbf{68.93\%}{\scriptsize\,($\uparrow$1.65)}
    & 2.80\%{\scriptsize\,($\uparrow$0.77)}
    & 5.47\%{\scriptsize\,($\uparrow$2.75)}
    & 53.52\%{\scriptsize\,($\uparrow$0.65)}
    & 33.87\%{\scriptsize\,($\downarrow$13.78)} \\

    & & & Masked
    & \textbf{63.73\%}{\scriptsize\,($\downarrow$3.55)}
    & 4.00\%{\scriptsize\,($\uparrow$1.97)}
    & 6.54\%{\scriptsize\,($\uparrow$3.82)}
    & 50.03\%{\scriptsize\,($\downarrow$2.84)}
    & 29.94\%{\scriptsize\,($\downarrow$17.71)} \\

    \cmidrule(lr){2-9}

    & \multirow{6}{*}{\textbf{TekGen}}
    & \multirow{3}{*}{\textbf{With}}
    & \cellcolor{green!20}Original
    & \cellcolor{green!20}26.68\%
    & \cellcolor{green!20}14.10\%
    & \cellcolor{green!20}28.16\%
    & \cellcolor{green!20}\textbf{36.95\%}
    & \cellcolor{green!20}21.13\% \\

    & & & Synthetic
    & 20.18\%{\scriptsize\,($\downarrow$6.50)}
    & 11.21\%{\scriptsize\,($\downarrow$2.89)}
    & 27.69\%{\scriptsize\,($\downarrow$0.47)}
    & \textbf{37.22\%}{\scriptsize\,($\uparrow$0.27)}
    & 20.54\%{\scriptsize\,($\downarrow$0.59)} \\

    & & & Masked
    & 25.33\%{\scriptsize\,($\downarrow$1.35)}
    & 10.07\%{\scriptsize\,($\downarrow$4.03)}
    & 19.98\%{\scriptsize\,($\downarrow$8.18)}
    & \textbf{31.16\%}{\scriptsize\,($\downarrow$5.79)}
    & 15.05\%{\scriptsize\,($\downarrow$6.08)} \\

    \cmidrule(lr){3-9}

    & &
    \multirow{3}{*}{\textbf{Without}}
    & \cellcolor{green!20}Original
    & \cellcolor{green!20}\textbf{36.82\%}
    & \cellcolor{green!20}5.86\%
    & \cellcolor{green!20}12.19\%
    & \cellcolor{green!20}27.59\%
    & \cellcolor{green!20}24.96\% \\

    & & & Synthetic
    & \textbf{35.57\%}{\scriptsize\,($\downarrow$1.25)}
    & 6.08\%{\scriptsize\,($\uparrow$0.22)}
    & 13.24\%{\scriptsize\,($\uparrow$1.05)}
    & 27.92\%{\scriptsize\,($\uparrow$0.33)}
    & 23.06\%{\scriptsize\,($\downarrow$1.90)} \\

    & & & Masked
    & \textbf{31.66\%}{\scriptsize\,($\downarrow$5.16)}
    & 5.59\%{\scriptsize\,($\downarrow$0.27)}
    & 11.64\%{\scriptsize\,($\downarrow$0.55)}
    & 23.07\%{\scriptsize\,($\downarrow$4.52)}
    & 18.82\%{\scriptsize\,($\downarrow$6.14)} \\

    \bottomrule
  \end{tabular}%
  }
\caption{Model comparison using macro-F1 for head-to-tail (HT) and
  tail-to-head (TH) on FewRel and TekGen. Bold indicates the best model
  per row, and rows with original entities are shaded.
  For synthetic and masked entities, arrows indicate the absolute
  macro-F1 change in percentage points relative to the corresponding
  original-entity result.}
  \label{tab:results}
\end{table}

For results without relation descriptions, Qwen3-4B achieves the highest macro-F1 score on FewRel (37.66\%) among the evaluated LLMs, whereas Flan-T5 XL achieves the highest macro-F1 score on TekGen (38.83\%) when using original entities in the head-to-tail multiple-choice question format. Llama-3.1-8B outperforms the other LLMs on both datasets using the tail-to-head multiple-choice question format. For synthetic entities, Flan-T5 XL achieves the highest macro-F1 scores on both datasets using the head-to-tail multiple-choice question format. However, Flan-T5 XL does not outperform Llama-3.1-8B when the latter uses the tail-to-head multiple-choice question format. Finally, we analyze the results obtained using masked entities in the question format. Consistent with the synthetic-entity setting, Flan-T5 XL achieves the highest performance on both datasets in the head-to-tail format, whereas Llama-3.1-8B achieves the highest performance in the tail-to-head format.

With relation descriptions, we conduct experiments using original, synthetic, and masked entities, as shown in \Cref{tab:results}. Using original entities, Llama-3.1-8B and Flan-T5 XL achieve the highest macro-F1 scores with the head-to-tail multiple-choice question format on FewRel (47.30\%) and TekGen (43.85\%), respectively. Using the tail-to-head format, Qwen2.5-7B and Qwen3-4B achieve the highest macro-F1 scores on FewRel (69.56\%) and TekGen (36.95\%), respectively. Using synthetic entities, Llama-3.1-8B and Flan-T5 XL obtain the highest macro-F1 scores with the head-to-tail format on FewRel (46.05\%) and TekGen (37.38\%), respectively, whereas Qwen3-4B achieves the highest performance with the tail-to-head format on both FewRel (58.60\%) and TekGen (37.22\%). Using masked entities, Flan-T5 XL achieves the highest macro-F1 scores on both FewRel and TekGen with the head-to-tail format, whereas Qwen3-4B consistently performs best on both datasets with the tail-to-head format. Overall, we cannot conclude that using original entities in the question format consistently outperforms using synthetic or masked entities.

In summary, adding relation descriptions to the head-to-tail prompt template increases the macro-F1 score for most LLMs on both datasets. However, we cannot conclude that adding relation descriptions consistently improves the recognition of inverse relations. Furthermore, no clear pattern emerges for the tail-to-head format. We further examine these results in terms of directional asymmetry, the impact of relation descriptions, and sensitivity to entity familiarity in \Cref{sec:discussion}.
\section{Discussion}
\label{sec:discussion}
\noindent This section discusses the experimental results on inverse relations from FewRel and TekGen. We analyze the results from three perspectives: (i) directional asymmetry, (ii) the impact of relation descriptions, and (iii) sensitivity to entity representations. To better interpret the results, we perform paired Wilcoxon signed-rank tests and report 95\% two-sided bootstrap confidence intervals computed from 10,000 bootstrap resamples. The reported 95\% confidence intervals correspond to the bootstrap confidence interval of the mean absolute paired difference and therefore quantify the magnitude of pairwise variability rather than the direction of the effect.

Each paired observation corresponds to the performance of one LLM on one inverse relation pair under one evaluation setting. The evaluation setting refers to either the prompting strategy (with or without relation descriptions) or the relation direction (head-to-tail or tail-to-head), depending on the statistical comparison. Consequently, the statistical tests are based on 70 paired observations for FewRel (7 inverse relation pairs × 2 evaluation settings × 5 LLMs) and 160 paired observations for TekGen (16 inverse relation pairs × 2 evaluation settings × 5 LLMs). TekGen's 16 inverse relation pairs contain negative relation labels, including those without any associated sentence in Appendix C.

\noindent\textbf{Directional Asymmetry in LLM Performance.} We evaluate relation classification by querying the relation between pairs of entities in both the head-to-tail and tail-to-head directions. To assess the effect of directionality, we perform paired Wilcoxon signed-rank tests, with the corresponding \emph{p}-values reported in \Cref{tab:head_tail_wilcoxon}. On FewRel, statistically significant performance differences are observed between the two directions for all entity types (\emph{p}<0.05). On TekGen, no statistically significant differences are observed for the original and synthetic entities, whereas a statistically significant difference is observed for the masked entities (\emph{p}<0.05). These results suggest that the effect of relation directionality depends on the dataset.
\begin{table}[H]
\centering

\resizebox{\columnwidth}{!}{%
\begin{tabular}{lllcccccccc}
\toprule
Dataset & Entity Type & Mean HT & Mean TH & Mean $|\Delta|$ &
95\% CI & Wilcoxon $W$ & $p$-value & $n$ \\
\midrule
FewRel & Original  & 35.0207 & 54.0912 & 30.0304 &
[24.6781, 35.6090] & 474  & 1.157e-05 & 70 \\

FewRel & Synthetic & 33.2289 & 49.3997 & 27.9056 &
[22.5095, 33.4868] & 557  & 0.0001005 & 70 \\

FewRel & Masked    & 34.5334 & 45.8181 & 30.0167 &
[24.6913, 35.6252] & 884  & 0.0359 & 70 \\
\midrule
TekGen & Original  & 56.4250 & 54.6105 & 43.1092 &
[38.3348, 48.0228] & 6098 & 0.7514 & 160 \\

TekGen & Synthetic & 56.9008 & 53.1041 & 42.6865 &
[38.0740, 47.2930] & 5771 & 0.3111 & 160 \\

TekGen & Masked    & 59.9211 & 47.0016 & 42.3880 &
[38.0605, 46.7261] & 4629 & 0.002913 & 160 \\
\bottomrule
\end{tabular}%
}
\caption{Paired Wilcoxon tests comparing results obtained using the head-to-tail (HT) and tail-to-head (TH) question formats, grouped by dataset and entity type. The 95\% confidence interval (CI) is reported for the mean absolute difference.}
\label{tab:head_tail_wilcoxon}

\end{table}
\noindent\textbf{Impact of Relation Descriptions.}
We further conduct statistical tests to assess the impact of including relation descriptions. As shown in \Cref{tab:description_effect}, including relation descriptions in the answer options leads to a marginal improvement for masked entities on FewRel (\emph{p}<0.1). In contrast, no statistically significant or marginal performance gains are observed for TekGen, as all corresponding \emph{p}-values exceed 0.1. These findings suggest that the impact of relation descriptions depends on the entity representation.
\begin{table}[H]
\centering

\resizebox{\columnwidth}{!}{%
\begin{tabular}{lllcccccccc}
\toprule
Dataset & Entity Type & Mean +D & Mean -D & Mean $|\Delta|$ &
95\% CI & Wilcoxon $W$ & $p$-value & $n$ \\
\midrule
FewRel & Original  & 46.6536 & 42.4583 & 17.6844 &
[14.2369, 21.4160] & 1002 & 0.1593  & 70 \\

FewRel & Synthetic & 42.9865 & 39.6420 & 18.0541 &
[14.6803, 21.6106] & 1029 & 0.2115  & 70 \\

FewRel & Masked    & 42.6194 & 37.7321 & 18.0436 &
[14.5568, 21.6465] & 936  & 0.07286 & 70 \\
\midrule
TekGen & Original  & 55.8641 & 55.1714 & 22.4289 &
[19.0480, 26.0239] & 4890 & 0.6031  & 160 \\

TekGen & Synthetic & 54.9810 & 55.0238 & 22.9345 &
[19.4702, 26.5241] & 5184 & 0.8304  & 160 \\

TekGen & Masked    & 54.1190 & 52.8038 & 23.3396 &
[20.1579, 26.5865] & 4861 & 0.6608  & 160 \\
\bottomrule
\end{tabular}%
}
\caption{Paired Wilcoxon tests comparing results obtained with relation descriptions (+D) and without relation descriptions (-D), grouped by dataset and entity type. The 95\% confidence interval (CI) is reported for the mean absolute difference.}
\label{tab:description_effect}
\end{table}
\begin{table}[H]
\centering
\resizebox{\columnwidth}{!}{%
\begin{tabular}{lllcccccccc}
\toprule
Dataset & Description & Comparison & Mean O & Mean Other &
Mean $|\Delta|$ & 95\% CI & Wilcoxon $W$ & $p$-value & $n$ \\
\midrule
FewRel & With D    & O vs.\ S & 46.6536 & 42.9865 & 6.7228  &
[5.4439, 8.0581] & 553.0  & 0.0002517 & 70 \\

FewRel & With D    & O vs.\ M & 46.6536 & 42.6194 & 11.0254 &
[9.1718, 12.9673] & 814.0  & 0.02826   & 70 \\

FewRel & Without D & O vs.\ S & 42.4583 & 39.6420 & 5.1150  &
[3.8960, 6.5112] & 587.0  & 0.0002073 & 70 \\

FewRel & Without D & O vs.\ M & 42.4583 & 37.7321 & 8.5054  &
[6.9266, 10.1533] & 606.0  & 0.0003227 & 70 \\
\midrule
TekGen & With D    & O vs.\ S & 55.8641 & 54.9810 & 4.1200  &
[3.2229, 5.1408] & 3492.0 & 0.05626   & 160 \\

TekGen & With D    & O vs.\ M & 55.8641 & 54.1190 & 13.9924 &
[11.4011, 16.8079] & 3985.5 & 0.06444   & 160 \\

TekGen & Without D & O vs.\ S & 55.1714 & 55.0238 & 2.8140  &
[2.2419, 3.4470] & 4240.0 & 0.73500   & 160 \\

TekGen & Without D & O vs.\ M & 55.1714 & 52.8038 & 11.8563 &
[9.5802, 14.4394] & 4104.5 & 0.06369   & 160 \\
\bottomrule
\end{tabular}%
}
\caption{Paired Wilcoxon tests comparing results obtained using original (O), synthetic (S), and masked (M) entities, with and without relation descriptions (D), by dataset. The 95\% confidence interval (CI) is reported for the mean absolute difference.}
\label{tab:entity_sensitivity}

\end{table}
\noindent\textbf{Sensitivity to Entity Representations.}
FewRel and TekGen are constructed from Wikidata. Because LLMs are trained on large-scale web data, they are likely to be familiar with at least some of these entities. To analyze the effect of entity familiarity, we replace the original entities with synthetic and masked entities. We then perform paired statistical tests, with the results reported in \Cref{tab:entity_sensitivity}. Statistically significant differences between the original and perturbed entities are observed for FewRel regardless of whether relation descriptions are included. For TekGen, marginal differences between the original and perturbed entities are observed when prompts include relation descriptions, whereas no statistically significant difference is observed between the original and synthetic entities without relation descriptions, as reported in \Cref{tab:entity_sensitivity}. These results suggest that LLMs exhibit sensitivity to entity representations, particularly on FewRel, although the effect is less consistent on TekGen. We also do not assess whether replacing original entities with synthetic entities changes sentence semantics or naturalness.

The statistical results reported in ~\Cref{tab:head_tail_wilcoxon}
highlight the impact of relation directionality on model performance for inverse relations in FewRel. On TekGen, no significant difference
is observed for original or synthetic entities, whereas a significant difference is found for masked entities. In particular, the findings suggest that LLMs often struggle to correctly classify inverse relations in FewRel. A further observation concerns benchmark datasets constructed from web-based resources, such as FewRel, as shown in \Cref{tab:entity_sensitivity}. Since LLMs are likely trained on large-scale web data, evaluations based solely on such resources may overestimate model generalization. This highlights the need for more robust evaluation strategies when benchmarking LLMs on datasets derived from web resources.

In conclusion, given that inverse relations are fundamental components of knowledge graphs, relation classification models should demonstrate robustness to directional variations. In this context, we assess the robustness of LLMs on inverse relations from the FewRel and TekGen datasets. Sensitivity to relation directionality is consistently observed on FewRel, whereas little or no significant performance difference is found for the original and synthetic entities on TekGen. These findings are consistent with the possibility that the evaluated LLMs benefit from prior exposure to entities or relation patterns
represented in FewRel and TekGen; however, the experiments do not directly establish memorization or dataset familiarity.

\section{Conclusion}
\label{sec:conclusion}
This work systematically evaluates five LLMs on inverse relations from FewRel and TekGen under multiple prompting settings. Specifically, we evaluate LLMs on inverse relations in two settings: with and without relation descriptions. Furthermore, we replace the original entities with synthetic and masked entities to assess robustness to entity anonymization.

Our findings show that adding relation descriptions does not significantly improve LLM performance on original entities (see \Cref{tab:description_effect}), except for inverse relations in FewRel evaluated with masked entities. Moreover, the inverse relation partition of FewRel is significantly affected by relation directionality, whereas the corresponding relations in TekGen are affected only when masked entities are used (see \Cref{tab:head_tail_wilcoxon}). Finally, we analyze the effect of entity familiarity by replacing the original entities with synthetic and masked entities (see \Cref{tab:entity_sensitivity}). The results in~\Cref{tab:entity_sensitivity} suggest that LLMs may be sensitive to changes in entity representations, particularly on FewRel, with weaker and less consistent evidence on TekGen.

In future work, we plan to investigate methods for disentangling entity familiarity from semantic changes introduced during anonymization and to evaluate inverse relation directionality on additional benchmarks.
\section*{Limitations}
\label{sec:limitations}
Our work has several limitations regarding entity anonymization and the evaluation of entity representations in LLMs.

First, we rely on an automated anonymization process that replaces entities known on the Web with alternative entities. However, we did not verify how many entities remained unchanged because they were not recognized by the anonymizer. We also did not assess whether the anonymization process altered the meaning or naturalness of the original sentences. In particular, relations that depend on entity attributes, such as \emph{father} and \emph{mother}, may be affected when the original entities are replaced. Consequently, the observed performance differences may reflect not only changes in entity representations but also changes in sentence semantics introduced by the anonymization process.

Second, this work does not introduce a new dataset but rather an evaluation benchmark derived from FewRel and TekGen. As a result, our findings are limited to these benchmark datasets and may not generalize to other relation classification benchmarks or knowledge graph domains. Furthermore, although the entity perturbation experiments provide insights into model sensitivity to different entity representations, they do not isolate the effect of entity familiarity from other factors introduced by the anonymization process. Therefore, our experiments should not be interpreted as direct evidence of memorization or of reliance on entity familiarity alone.

Lastly, the language of these data from FewRel and TekGen is English; therefore, our evaluation cannot be generalized to other languages.
\section*{Ethical Considerations}
The benchmark is derived from the publicly available FewRel and TekGen datasets, both of which are distributed under the Creative Commons Attribution-ShareAlike 4.0 International (CC BY-SA 4.0) license.


\bibliographystyle{splncs04}
\bibliography{sample-base}

\appendix

\appendix
\section*{A.) Declaration of Generative AI Use}
\noindent During the preparation of this work, a generative AI model (GPT 5.1 with a temporary chat option) was used to check grammar and spelling. After using this tool/service, the authors reviewed and edited the content as needed and took full responsibility for the publication's content.
\section*{B.) Sample From FewRel Dataset}
We present a sample from FewRel below.
\begin{tcolorbox}[
  title={Example Data of Inverse Relations From FewRel.},
  breakable,
  colback=gray!3,
  colframe=black!80,
  boxrule=0.5pt
]
\begin{lstlisting}[
  basicstyle=\ttfamily\scriptsize,
  breaklines=true,
  numbers=left,
  numbersep=6pt,
  frame=none,
  columns=fullflexible,
  keepspaces=true
]
{
  "head_to_tail": "P40",
  "tail_to_head": "P22",
  "tokens": ["Niels", "Bohr", "and", "his", "son",
             "Aage", ",", "a", "physicist", "who", "acted",
             "as", "his", "father", "'s", "assistant", ",",
             "arrived", "on", "30", "December", "on", "the",
             "first", "of", "several", "visits", "as", "a",
             "consultant", "."],

  "tail": ["niels bohr", "Q7085", [[0, 1]]],
  "head": ["aage", "Q103854", [[5]]]
}
\end{lstlisting}
\end{tcolorbox}

\section*{C. List of All Inverse Relations in TekGen}
\Cref{tab:inverse_properties} presents all inverse relation pairs in TekGen, including the relations without a sentence.
\begin{table*}[t]
\centering
\small
\begin{tabularx}{\textwidth}{@{}lXlX@{}}
\toprule
\textbf{Property} & \textbf{Description} &
\textbf{Inverse} & \textbf{Inverse Description} \\
\midrule
Astronaut mission &
Space mission that the subject is or has been a member of (excluding future missions). &
Crew member &
Person that participated operating or serving aboard this vehicle. \\

Taxon synonym &
Name listed as a synonym of this item's taxon name. &
Taxon synonym of &
Preferred taxon name of which this item's taxon name is a synonym. \\

Characters &
Characters appearing in books, films, TV series, comics, operas, or video games. &
Present in work &
Entity appearing in a work as part of its narration. \\

Creator &
Maker of a creative work or object when no more specific property exists. &
Notable work &
A notable scientific, artistic, literary, or other significant work created by the subject. \\

Coach of sports team &
Sports club or team for which the person served as coach or manager. &
Head coach &
On-field manager or head coach of a sports club or person. \\

Depicts &
Entity visually depicted, described, or otherwise represented in a creative work. &
Depicted by &
Creative work that depicts this subject. \\

Part of &
Object of which the subject is a part. &
Has part &
Component that forms part of the subject. \\

Position held &
Position or public office currently or formerly held by the subject. &
Officeholder &
Person who currently or previously held the position. \\

Cast member &
Actor appearing in a production. &
Character role &
Role or character portrayed by the cast member. \\

Followed by &
Immediately succeeding item in a series. &
Follows &
Immediately preceding item in a series. \\

Based on &
Work or input used as the basis of the subject item. &
Derivative work &
Work derived from a substantial part of this work. \\

Commanded by &
Commander of a military unit, operation, or organization. &
Commander of &
Military units or organizations commanded by the person. \\

Reservoir created &
Reservoir created upstream by a dam. &
Dam &
Construction that impounds the watercourse or creates the reservoir. \\

Habitat &
Natural environment in which an organism or species lives. &
Taxon found at location &
Taxon documented as occurring at a specific location. \\
\bottomrule
\end{tabularx}
\caption{All inverse relation pairs in TekGen}
\label{tab:inverse_properties}
\end{table*}
\end{document}